\documentclass[preprint,12pt]{elsarticle}

\usepackage{amsmath,amssymb,amsthm}
\usepackage{graphicx}
\usepackage{booktabs}
\usepackage{multirow}
\usepackage{algorithm}
\usepackage{algpseudocode}
\usepackage{hyperref}
\usepackage{url}
\usepackage{subcaption}
\usepackage{float}

\begin{document}
\begin{frontmatter}
\title{LLM Inference Under Bursty Workload Distribution: Modifying the WAIT Algorithm}

\author[inst1]{Anjali Gangadhar Katageri}
\ead{anjalik181004@gmail.com}

\author[inst2]{Shobha Rani\corref{cor1}}
\ead{srani1@ma.iitr.ac.in}

\author[inst2]{Raghu Nandan Sengupta}
\ead{raghus@iitk.ac.in}

\affiliation[inst1]{organization={Department of Mathematics, Indian Institute of Technology Dharwad},
city={Dharwad}, state={Karnataka}, country={India}}

\affiliation[inst2]{organization={Department of Management Sciences, Indian Institute of Technology Kanpur},
city={Kanpur}, state={Uttar Pradesh}, country={India}}

\cortext[cor1]{Corresponding author}

\begin{abstract}
Large Language Models (LLMs) such as ChatGPT and Claude are widely used for information retrieval and problem-solving. Recent work has focused on improving scheduling algorithms to boost throughput while maintaining low latency. However, these approaches often assume Poisson request arrivals with constant rates—an assumption that fails to reflect the inherently bursty and  dynamic nature of real-world traffic.

We propose a lightweight extension to the state-of-the-art WAIT algorithm~\cite{ao2025optimizing}, which adapts to time-varying arrival rates without prior traffic knowledge. The proposed algorithm performs online estimation of request intensity based on observed interarrival times. Using Markov Modulated Poisson Process (MMPP) based synthetic workloads with diverse request types, we conduct a simulation-based evaluation demonstrating that the proposed method achieves higher throughput than Sarathi-Serve~\cite{agrawal2023sarathi}, ORCA~\cite{yu2022orca}, and vLLM~\cite{kwon2023efficient} in the evaluated low arrival-rate shift scenarios while maintaining comparable latency. Source code for our experiments is available at \url{https://github.com/anjalik04/modified_wait}.

\end{abstract}
\end{frontmatter}

\section{Introduction}
%
%
%
%
Since the public release of ChatGPT in 2022, Large Language Models (LLMs) have become integral to applications such as education, healthcare, customer support, and software development. Models like ChatGPT, Claude, Gemini, LLaMA, and Mistral have increased demand for low-latency inference, creating challenges in scaling LLM-serving systems under GPU memory and compute constraints.

Recent frameworks such as vLLM, FasterTransformer, and DeepSpeed-MII improve throughput through techniques like KV-cache reuse, memory paging, and batching. However, they are typically evaluated under fixed-rate arrivals, which do not reflect the bursty and time-varying traffic patterns observed in production systems. Prior work, including BurstGPT and ELIS, shows that real-world LLM workloads often deviate significantly from these assumptions.

To address this gap, we propose a modification to the WAIT scheduling algorithm that incorporates online rate estimation to better handle bursty, non-stationary workloads. Using Markov Modulated Poisson Processes (MMPP) to model dynamic arrivals, we demonstrate through simulation that our approach achieves competitive throughput and latency across bursty workloads. In the evaluated low arrival-rate shift scenarios, the proposed method achieves higher throughput than Sarathi, ORCA, and vLLM while maintaining comparable latency.
 The contributions of this paper are as follows:
\begin{itemize}
    \item We propose a lightweight modification to the WAIT scheduling algorithm (proposed algorithm) by introducing an online arrival rate estimation mechanism that adapts to bursty traffic patterns.
    \item To capture realistic temporal variation in LLM request patterns, traffic is modeled using a two-state Markov Modulated Poisson Process (MMPP-2).
    \item A comprehensive simulation framework is utilized to evaluate Sarathi, ORCA, vLLM, and both original and modified WAIT algorithms under varying traffic conditions.
    \item We empirically demonstrate that the Modified WAIT algorithm achieves throughput comparable to, or better than, state-of-the-art systems in scenarios characterized by moderate shifts in arrival rates (`Low Shift' conditions).
\end{itemize}

The rest of this paper is structured as follows. We begin with the technical background on LLMs and the motivation behind this work in Section II, followed by a review of related literature in Section III. Section IV presents the proposed Modified WAIT algorithm, while Section V outlines the methodology used. Performance evaluation results are discussed in Section VI. Finally, Section VII concludes the paper and highlights future research directions.

\section{Background and Motivation}

We first discuss LLM inference, performance metrics, request scheduling, and workload distribution in large language model serving systems.

\subsection{{LLM Inference}}
\begin{figure}
    \centering
    \includegraphics[width=0.5\linewidth]{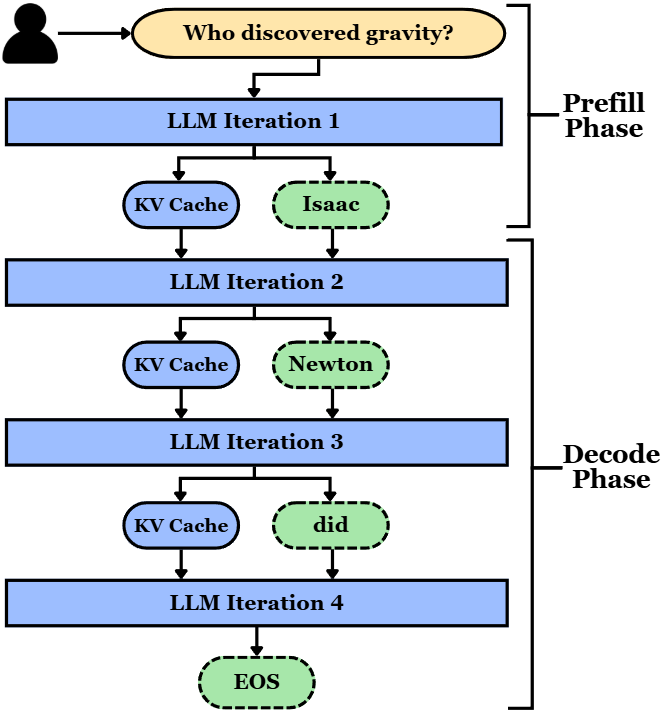}
    \caption{Flowchart representing prefill and decode phases of LLM inference}
    \label{fig:llm_inference_flowchart}
\end{figure}

LLMs generate text in an autoregressive manner, producing one token at a time conditioned on prior context. This process known as inference unfolds in two stages: the \textbf{prefill phase} and the \textbf{decode phase}. During prefill, the full prompt is tokenized, embedded, and passed through the transformer layers, generating key value (KV) pairs stored in a KV cache for future use. This stage supports full parallelization, enabling efficient processing of long inputs. In the decode phase, the model sequentially generates one token at a time by embedding the previously generated token and processing it through the transformer using the pre-computed KV cache. Only the new KV for the current token needs to be calculated, reducing redundant computation. At each step, a probability distribution over the vocabulary is produced until a termination condition such as maximum length or end of sequence (EoS) token is met. Figure~\ref{fig:llm_inference_flowchart} describes pictorially the flow process of prefill and decode phases of LLM inference. Despite its sequential nature, the decode phase remains efficient due to the reuse of cached attention values. 
 
Recent efforts to optimize LLM inference have focused on reducing latency, memory usage, and improving throughput through techniques such as dynamic batching, KV-cache reuse, and distributed execution. Studies \cite{agrawal2023sarathi, chen2024punica, zhong2024distserve} have investigated batching strategies and their impact on performance, while works like \cite{vllm2023, sheng2023faster, zheng2024smoothquant, rajbhandari2022deepspeed, anil2023palm} demonstrate significant improvements across both prefill and decode stages, enabling more efficient LLM serving.

\subsection{{Performance Measures}}
The performance of an LLM inference system is an important factor and assessed using several key metrics, including:
\begin{enumerate}
\item[(i)] \textbf{Latency:} The average time taken from when a user submits a request to when the final response is delivered. In practice, both mean latency and tail latency (e.g., 95th or 99th percentile) are important. High latency leads to poor user experience, particularly in interactive settings.

\item[(ii)] \textbf{Throughput:} Defined as the number of tokens generated per second, reflects the model's overall computational efficiency. Higher throughput means that more users can be served concurrently.

\item[(iii)] \textbf{Time To First Token (TTFT):} This metric quantifies the delay between a request's arrival and the generation of the initial token. It is critical for perceived responsiveness, particularly in interactive applications.

\item[(iv)] \textbf{Fairness and Starvation:} In a multi-user setting, fairness metrics are important to ensure that no request is indefinitely delayed. Starvation, wherein long requests are continuously bypassed in favor of shorter ones, can severely degrade user experience if not addressed by scheduling policy design.

\end{enumerate}

Several studies have highlighted the trade-offs between latency, throughput, and fairness in LLM serving systems (\cite{vllm2023, sheng2023faster, rajbhandari2022deepspeed}). Techniques such as adaptive scheduling, dynamic batching, and quality-of-service (QoS) aware routing have been proposed to strike a balance between minimizing latency and maximizing resource utilization \cite{anil2023palm}. These works enables us to understand how the design of practical inference infrastructure that meets diverse performance goals in real-time deployments can be done.

\subsection{Request Scheduling}
Request scheduling is central to the performance of LLM inference systems, directly impacting latency, throughput, memory efficiency, and fairness. Simple policies like First-Come-First-Serve (FCFS) can degrade performance by allowing long requests to block shorter ones, increasing mean and tail latencies. In contrast, strategies such as Shortest Job First (SJF) prioritize short prompts and improve responsiveness. However, in transformer-based models that rely on KV-cache for sequential decoding, scheduling decisions also affect memory usage and cache efficiency. Interrupting long-running requests for shorter ones may lead to cache evictions, risking memory overload. On a larger scale, particularly in environments involving multiple tenants, scheduling balances responsiveness, fairness, and resource utilization, making it a critical component of high-performance LLM service.



Furthermore, designing efficient scheduling policies requires careful attention to advanced system behaviors, such as preemption and their unintended consequences like starvation. 
In parallel, formal analysis of scheduling strategies is greatly enriched by queueing-theoretic frameworks for modeling and optimizing inference performance under diverse workloads.

\subsubsection{\textbf{Fixed Arrival Rates}}
A fixed-rate arrival process is commonly modeled as a Poisson process, \( X \sim \text{P}(\lambda)\), where inter-arrival times are exponentially distributed with rate $\lambda$. Due to the memoryless property, the system is well-represented by the M/M/1 queue—Memoryless arrivals, Memoryless service, and one server enabling tractable analysis through closed-form performance metrics.

\subsubsection{\textbf{Bursty Workload Distribution}}
In practical LLM deployments, user request arrivals deviate significantly from the constant rate assumption of Poisson models. Instead, traffic exhibits \textbf{burstiness} characterized by sudden spikes in request volume followed by lulls driven by human interaction patterns in applications such as chatbots and content generation. Empirical studies (\cite{2401.17644, 2505.09142}) based on production traces confirm this behavior, revealing substantial deviations from stationarity and memorylessness. Consequently, classical queueing models with exponential inter-arrival assumptions often fail to capture the performance dynamics of real world systems.
\subsection{Markov-Modulated Poisson Process (MMPP)}
MMPP is a well-established stochastic process for modeling bursty and correlated arrival processes. Our objective is not to reproduce a specific production workload but to evaluate scheduling behaviour under controlled levels of burstiness. Unlike a basic Poisson process with a fixed arrival rate, an MMPP allows the rate to switch over time, governed by a finite state continuous time Markov chain. Each state represents a different traffic intensity, with its own Poisson arrival rate (e.g., 10 vs. 100 requests/sec). The system randomly transitions between these states, remaining in each state for an exponentially distributed duration. This structure introduces temporal correlation and sustained high or low traffic periods, effectively capturing real world usage spikes driven by time of day, events, or human interaction. Formally, the MMPP is defined by two components:
\begin{enumerate}
    \renewcommand{\labelenumi}{\roman{enumi}.}
    \item A \textbf{generator matrix} that determines the transition rates between states in the underlying Markov process.
    \item A vector of \textbf{Poisson rates}, where each element corresponds to the request arrival rate in one of the states.
\end{enumerate}

By adjusting the transition rates and arrival intensities across states, MMPPs can simulate various burstiness levels. Rapid state transitions lead to frequent fluctuations in request arrivals, while slower transitions create prolonged bursts or idle periods mimicking patterns like diurnal cycles. This dual ability to capture both short-term variability and long-term trends 
\begin{figure}
    \centering
    \includegraphics[width=0.8\linewidth]{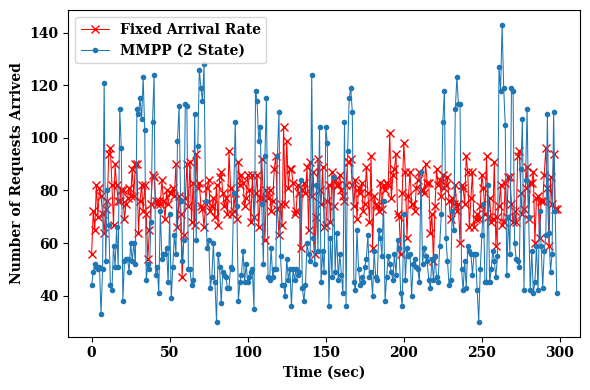}
    \caption{Comparison of Request Arrival: Fixed rate Poisson vs. MMPP (2-State)}
    \label{fig:placeholder}
\end{figure}
makes MMPP a powerful tool for modeling dynamic workloads. It provides a rigorous framework to evaluate how well scheduling algorithms perform under real world, non-stationary traffic ensuring system robustness even during demand spikes.

\begin{table}[H]
\centering
\caption{Evidence supporting the use of MMPP-2 for modeling bursty LLM inference workloads.}
\label{tab:mmpp_validation}
\small
\begin{tabular}{p{3.2cm} p{5.2cm} p{5.2cm}}
\toprule
\textbf{Application Domain} &
\textbf{Bursty Behavior Captured by MMPP-2} &
\textbf{Relevance to LLM Serving} \\
\midrule

Teletraffic \cite{dong2017copula}
&
Alternating periods of high and low call arrival intensity
&
Diurnal fluctuations in prompt arrival rates
\\

Web Services \cite{casale2012dealing}
&
Flash crowds interleaved with idle periods
&
Sudden surges caused by trending events or viral content
\\

IoT / M2M Networks \cite{elfawal2024markov}
&
Intermittent transmissions separated by inactivity
&
Session-oriented user activity and request clustering
\\

Cloud Transaction Systems \cite{yoshihara2001practical}
&
Long-range burstiness and correlated arrivals
&
Sustained conversational workloads generating correlated requests
\\

LLM Production Traces \cite{wang2025burstgpt,choi2025elis}
&
High inter-arrival variability ($\mathrm{CV}>2.5$) and clear high/low demand regimes
&
Observed traffic naturally exhibits bursty state transitions consistent with MMPP-2
\\

\bottomrule
\end{tabular}
\end{table}

Figure~\ref{fig:placeholder} compares request arrival traces generated by two modeling approaches: a fixed rate Poisson process and an MMPP. The MMPP alternates between two states with arrival rates of 50 and 120 requests per second. For fairness, the Poisson process uses a constant rate equal to the time averaged mean of the MMPP, ensuring both traces produce the same total number of arrivals over the simulation period. Despite this, their temporal patterns differ markedly: the MMPP trace displays clear burstiness, with sustained high and low traffic periods that closely mirror real world workloads. In contrast, the Poisson process produces a more uniform and memoryless arrival pattern, failing to capture the intermittent surges in traffic that are typical in practical systems.

MMPP-2 has been widely used to model bursty arrival processes in telecommunication, web, IoT, and cloud systems. As shown in Table~\ref{tab:mmpp_validation}, the traffic characteristics observed in these domains closely resemble those reported in recent LLM serving traces, including alternating high- and low-demand periods, clustered arrivals, and high variability in inter-arrival times. These similarities motivate the use of MMPP-2 as a parsimonious yet effective model for bursty LLM workloads.
\subsection{Problem Statement}
The inference workloads of LLMs in real-world applications are characterized by non-stationary, bursty traffic patterns that vary over time. Existing scheduling algorithms such as WAIT have been shown to perform well under idealized conditions with known and stationary request arrival rates.

However, these algorithms depend on prior knowledge of the full arrival trace to compute optimal booking limits, making them unsuitable for deployment in live systems where future request patterns are unknown and highly dynamic. This creates a fundamental limitation in their applicability. Therefore, there is a critical need to develop a scheduling algorithm that retains the theoretical strengths of WAIT while dynamically adapting to real-time bursty workloads without relying on any pre-observed data trace.

\section{Literature Review}
LLM inference introduces several complexities that challenge traditional scheduling methods. Most notably, prompt arrivals are stochastic and bursty, and inference itself comprises multiple distinct phases—a prefill stage followed by token-wise autoregressive decoding. Classical operations research techniques, which often rely on assumptions of steady-state behavior or single-phase service, fall short in this setting. The dynamic memory footprint and computational heterogeneity of LLM workloads, especially the growth of KV-cache during decoding, necessitate new scheduling strategies tailored to these evolving demands. Failing to manage memory properly may cause KV-cache to forcefully switch to slower storage tiers, significantly increasing latency and reducing throughput~\cite{tay2022efficient, kang2024gear,hooper2024kvquant}.

To enhance GPU utilization and throughput, current LLM-serving frameworks typically batch multiple prompts and coordinate execution across the prefill and decode phases. The evolution of models such as GPT-3 and GPT-4~\cite{brown2020language,chowdhery2023palm,kaplan2020scaling,openai2023gpt} has propelled impressive gains in generative capabilities, yet many of the corresponding optimizations—such as batching heuristics, model parallelism, and KV-cache compression~\cite{kwon2023efficient,yu2022orca,liu2024deepseek,zhong2024distserve}—remain engineering-centric and lack a strong theoretical foundation. This limits their generalizability, especially in non-stationary environments with fluctuating request patterns.

A key bottleneck in inference scheduling is the unpredictability of output token lengths. While some systems attempt to estimate this length in advance, the predictions are often imprecise and highly sensitive to classification granularity~\cite{fu2024efficient}. As a result, forecast-driven scheduling is susceptible to errors and often unreliable.

Recognizing the limitations of static scheduling approaches, several recent works have explored online and adaptive scheduling algorithms. These techniques aim to learn patterns from data and adjust policies accordingly. For instance, \cite{cole2014sample,balkanski2016power, lattanzi2020online} explore strategies that predict arrival distributions to improve job scheduling over time, while \cite{li2020online} specifically focuses on joint optimization of batching and scheduling via job clustering. When no prior knowledge of workload patterns is available, such methods may suffer from degraded performance due to uninformed decisions.

A significant challenge in LLM-serving systems is burstiness, where traffic can fluctuate dramatically over short periods. This behavior, well-documented in networked and service-oriented systems~\cite{mi2007performance}, is difficult to model with classical Poisson-based assumptions. Instead, Markovian models such as the Markov Arrival Process (MAP) and its widely used subclass, the MMPP, offer a more expressive framework \cite{fischer1993markov} for capturing time-varying request rates. MMPP-based models have been successfully applied to simulate bursty LLM traffic\cite{casale2012dealing,horvath2002markovian,okamura2009faster}, and parameter fitting methods based on real traffic traces have been developed for two-state models~\cite{perez2011performance,perez2012analysis} to represent normal and high-load modes.A recent study \cite{mitzenmacher2025queueing} identifies key gaps in applying queueing theory to LLMs and proposes new research directions grounded in predictive scheduling and stochastic modeling.

Despite these challenges, much of the existing work remains system-oriented, focusing on engineering solutions rather than principled theory. For instance, Splitwise~\cite{patel2023peeling} and DistServe~\cite{zhong2024distserve} decompose LLM inference into distinct processing stages to better manage system resources. Meanwhile, Sarathi~\cite{agrawal2023sarathi} and ORCA~\cite{yu2022orca} implement batching-based strategies to boost throughput and latency trade-offs. These efforts are effective but often lack formal guarantees.

Encouragingly, recent theoretical efforts have begun to close this gap. Notably,~\cite{li2025throughput} proposes a stochastic model for LLM inference that assumes a linear dependency between batch size and processing time. Their results show that work-conserving scheduling policies—which ensure that resources are always utilized when jobs are present—can achieve optimal throughput in both single-user and multi-agent LLM settings.

\section{Modifying the WAIT Algorithm}
The WAIT algorithm addresses scheduling challenges in memory-constrained LLM inference systems. To contextualize our modifications, we first provide a concise overview of WAIT’s core mechanisms and limitations, followed by our enhancements for dynamic workloads.

\subsection{Prompts and Inference Process}
This subsection describes the prompt characteristics then explain the batching and GPU processing dynamics proposed by the WAIT algorithm.
\subsubsection{\textbf{Prompts Characteristics}}
Prompts arrive stochastically and are classified into $m$ distinct types indexed by $j \in \{1, \dots, m\}$. Each request type represents a category of inference queries with similar input and output characteristics, such as comparable prompt lengths and expected generation lengths. We assumed that the prompts of type $j$ arrive according to a Poisson process with rate $\lambda_j$. Each type $j$ prompt is characterized by two primary parameters:
\begin{enumerate}
    \renewcommand{\labelenumi}{\roman{enumi}.}
    \item an input prompt length of $l_j$ tokens after tokenization and prefill 
    \item a total sequence length of $l_j + l'_j$ tokens after completion of decoding, where $l'_j$ denotes the number of output tokens generated by the model
\end{enumerate}

Thus, a type-$j$ prompt undergoes one prefill iteration followed by $l'_j$ decode iterations to complete its inference processing. During the prefill phase, the prompt’s input context and question are embedded as $l_j$ tokens in a single iteration, consuming a KV cache of size $l_j$ units. In the subsequent decode phase, each output token is generated per iteration, adding one KV cache unit. A prompt is in \textbf{stage 0} before prefill, and in \textbf{stage $k$} during its $k$-th decode iteration.

\subsubsection{\textbf{Batching and GPU Processing Dynamics}}
To maximize GPU efficiency, tasks are processed in \textit{batches}, combining prompts for prefilling and prefilled prompts for decoding. Consider a batch consists of:
\begin{itemize}
    \item $n_1$ prompts in the prefill phase, each with an input length of $k_i$ tokens for $i = 1, 2, \dots, n_1$. These prompts are processed to initialize their KV caches.
    \item $n_2$ prompts in the decode phase, each with output length of $s_i = l_{j(i)} + k_i$ tokens after the current iteration, where $l_{j(i)}$ is the input length of prompt $i$ of type $j(i)$, and $1 \leq k_i \leq l'_j$ denotes the number of tokens generated so far in the decode phase for prompt $i$.
\end{itemize}

\subsection{WAIT Scheduling Algorithm}
The \textit{Waiting for Accumulated Inference Threshold} (WAIT) algorithm is a scheduling policy designed to maximize throughput in LLM inference systems under known prompt arrival types, while adhering to memory constraints and performance metrics including latency and time-to-first-token (TTFT). Leveraging the fluid dynamics established from the field of Queueing Theory, WAIT tries to optimize batch formation and scheduling by accumulating prompts until specific thresholds are met, ensuring efficient resource utilization in stochastic environments. The algorithm assumes that memory capacity $C \geq M^*$, the equilibrium memory capacity.

The WAIT algorithm operates by maintaining $m$ inventories of prompts and tracking their current inventory across all types and stages until the system reaches a state in accordance with the fluid dynamics. For each prompt type $j \in [m]$ and stage $k \in \{0, 1, \dots, l'_j\}$, the algorithm sets a threshold $n_j$ for type $j$ prompts, where $n_j$ is derived from the fluid dynamics and will be detailed subsequently.

At each continuous time $t$, the WAIT algorithm monitors the current inventory $n^t_{jk}$, defined as the number of waiting prompts of type $j$ at stage $k$, for $k = 0, 1, \dots, l'_j$ , alongside stochastic arrivals following Poisson distribution i.e. \( X \sim \text{Poisson}(\lambda_j) \) in each unit time interval. At time $t$, the algorithm incorporates type $j$ prompts into the batch for a new iteration whenever the inventory level satisfies:$$n^t_{j0} \geq n_j.$$ 
This condition ensures that type $j$ prompts are processed at time $t$ only if at least $n_j$ such prompts are waiting at the prefill stage. When satisfied, WAIT constructs a batch $B^t$ by selecting $\min(n_j, n^t_{jk})$ type $j$ prompts at each stage $k = 0, 1, \dots, l'_j$, for every $j$ meeting the threshold. This batch is processed in the current iteration, utilizing GPU memory proportional to the total number of token processed. After processing, the inventory is updated and prompts advanced stages. If no type $j$ satisfies its threshold, the algorithm pauses to accumulate more prompts. Moreover, the algorithm preserves KV caches computed for waiting prompts to avoid recomputation. This threshold-based batching aligns with fluid dynamics, striving to minimize memory waste and enhance throughput under heavy load. The WAIT algorithm is detailed in Algorithm~\ref{alg:wait}.

\begin{algorithm}
\small
\caption{WAIT: Waiting for Accumulated Inference Threshold}
\label{alg:wait}
\begin{algorithmic}[1]
\Require Memory $C$, arrival rates $\lambda_j$, thresholds $n_j$ satisfying (\ref{threshold_eq_wait}), $\forall j \in [m]$
\State Initialize prompt inventory $n_{jk} \gets 0$ for all $j \in [m], k \in \{0, 1, \dots, l_j'\}$
\State Initialize event queue with arrival events for each type $j$
\State Set current time $t \gets 0$
\While{True}
    \State Wait for the next event (arrival or batch completion)
    \If{event is an arrival of type $j$}
        \State Update inventory: $n_{j0} \gets n_{j0} + 1$ \Comment{Add new prompt to waiting queue of prefill phase}
    \ElsIf{event is a batch completion}
        \State Update inventory: For each $j$ in batch $B$, $n_{jk} \gets n_{jk} - \min(n_j, n_{jk})$ for $k = 0, \dots, l_j'$
        \State Advance prompts: For each $j$, move $\min(n_j, n_{jk})$ prompts from stage $k$ to $k+1$
        \State Clear KV caches for prompts reaching stage $l_j' + 1$ \Comment{Free memory for completed prompts}
    \EndIf
    \State Check if $n_{j0} \geq n_j$ for any $j \in [m]$ \Comment{Check threshold for batching}
    \If{condition $n_{j0} \geq n_j$ is met for some $j$}
        \State Form batch $B$ by selecting $\min(n_j, n_{jk})$ prompts of type $j$ at each stage $k$ for all $j$ meeting the condition \Comment{Create batch}
        \State Process batch $B$, keep prompts outside the batch waiting and do not delete their KV caches
    \EndIf
\EndWhile
\end{algorithmic}
\end{algorithm}

When the memory constraint satisfies \( C \geq M^* \), consider the WAIT policy \( \pi \), parameterized by \( [n_1, \dots, n_m] \), where \( n_j \) denotes the threshold number of type-\( j \) prompts at each stage. This policy is asymptotically optimal when the thresholds satisfy the following conditions:

\begin{equation}
\begin{split}
\Delta T(n_{1:m}) &:= d_0 + d_1M^\pi \leq \frac{n_j}{\lambda_j}, \quad \forall j \in [m], \\
M^\pi &= \sum_{j=1}^{m} n_j(l'_j + 1)\left(l_j + \frac{l'_j}{2}\right)
\end{split}
\label{threshold_eq_wait}
\end{equation}

\subsection{Developing an Intuition for Modification}
Our objective is to improve the WAIT algorithm's effectiveness under bursty traffic modeled by an MMPP. Recognizing that MMPP arrival rates shift across states over time, we initially aimed to adapt the scheduler's thresholds accordingly. However, such abrupt adjustments to arrival rates led to unstable decoding cadence and erratic memory usage, significantly degrading performance. Interestingly, we found that the original WAIT policy performed reasonably well when provided with a single, global mean arrival rate accross all request types. This raised a key design choice: whether to update thresholds after each request or in batches. Per-request updates offer responsiveness but incur higher computation, while batch updates (e.g., every 10 requests) may delay adaptation. Experiments showed minimal performance difference between frequent and infrequent threshold updates, so we adopt the simpler per-request update, computing the moving average after each arrival.

\section{Methodology}
This section describes the methodology used to develop the Modified WAIT algorithm. We first generate synthetic bursty request traces using an MMPP-2 model, then estimate arrival rates online from the observed request stream, and finally use these estimates to dynamically update the batching thresholds of the original WAIT algorithm.
\subsection{Synthetic Trace Description}
To simulate bursty workloads resembling real-world traffic, we adopt the MMPP framework. Among the various forms of MMPP, we employ the two-state MMPP (MMPP-2). Several MMPP-2 models have been widely adopted in teletraffic and network traffic modeling because they capture bursty behavior using just two load states while permitting efficient parameter estimation ~\cite{Rydén01011994}, \cite{2401.14561}. Its analytical tractability also enables closed-form derivations of key performance metrics, such as effective bandwidth and queue lengths~\cite{app14188561, doiPracticalTimeScale}. Thus, MMPP-2 remains a principled and widely used model in performance analysis. The computational steps are provided in Algorithm~\ref{alg:mmpp} as:

\begin{algorithm}
\small
\caption{Generating 2-State MMPP Trace}
\label{alg:mmpp}
\begin{algorithmic}[1]
\Require Total time $\text{T}_{\max}$, Poisson rates $\lambda_0$, $\lambda_1$
\Require Transition rates $q_{01}$ (from state 0 to 1), $q_{10}$ (from state 1 to 0)
\Ensure List of arrival timestamps and state intervals

\State Initialize empty lists $\displaystyle{arrivals}$, $\displaystyle{state\_times}$
\State Set $\displaystyle{current\_time} \gets 0$
\State Randomly choose $\displaystyle{state} \in \{0, 1\}$

\While{$\displaystyle{current\_time} < \text{T}_{\max}$}
    \If{$\displaystyle{state} = 0$}
        \State $\displaystyle{holding\_time} \gets \text{Exponential}(1 / q_{01})$
        \State $\displaystyle{rate} \gets \lambda_0$
    \Else
        \State $\displaystyle{holding\_time} \gets \text{Exponential}(1 / q_{10})$
        \State $\displaystyle{rate} \gets \lambda_1$
    \EndIf
    \State $\displaystyle{end\_time} \gets \min(\displaystyle{current\_time} + \displaystyle{holding\_time}, \text{T}_{\max})$
    \State $\displaystyle{duration} \gets \displaystyle{end\_time} - \displaystyle{current\_time}$
    \State $\displaystyle{num\_arrivals} \gets \text{Poisson}(\displaystyle{rate} \times \displaystyle{duration})$
    \State Generate $\displaystyle{num\_arrivals}$ uniform timestamps in $\displaystyle{[{current\_time}, {end\_time}]}$
    \State Append sorted timestamps to $\displaystyle{{arrivals}}$
    \State Append $\displaystyle{({current\_time}, {end\_time}, {state}, {rate})}$ to $\displaystyle{{state\_times}}$
    \State $\displaystyle{{current\_time} \gets {end\_time}}$
    \State $\displaystyle{{state} \gets 1 - {state}}$ \Comment{Toggle state}
\EndWhile

\State \Return $\displaystyle{{arrivals}, {state\_times}}$
\end{algorithmic}
\end{algorithm}

In the MMPP-2 setup, the system switches between two latent states, each with a distinct Poisson arrival rate. Transitions between these states occur probabilistically, with the time spend in each state being exponentially distributed. One state represents a low-load regime, while the other captures high-load bursts with elevated arrival rates.

Although higher-order MMPP models can represent more intricate traffic patterns (e.g., varying burst intensities or periodicities), we defer their exploration to future work to preserve clarity and computational tractability in this study.

\subsection{Estimating Arrival Rates: Single Request Type}
We begin by describing our method in the context of a single request type. The extension of this method to multiple request types will be discussed in subsequent sections.


\subsubsection{\textbf{Preprocessing and Smoothing of Arrival Traces}}
At runtime, the system observes only arrival timestamps and prompt tokens of incoming requests. From these, we compute interarrival times as the basis for estimating arrival rates. To reduce noise and improve stability, we preprocess the interval arrival time series by appying a smoothing operation. Specifically, for a window size $w$, we compute the average of the most recent $w$ interarrival times. The local arrival rate $\lambda_t$ is then approximated as:
\[
\lambda_t = \frac{1}{\text{interarrival}_t}
\]
Due to burstiness, these rates fluctuate over time. To extract underlying trends, we apply an Exponential Moving Average (EMA), which gives more weight to recent values while retaining long term trends, defined as:
\[
EMA_t = \alpha \cdot r_t + (1 - \alpha) \cdot EMA_{t-1},
\]
where $\alpha \in (0,1)$ is a smoothing factor that controls the rate of decay. This is further refined with a Savitzky Golay filter, which fits low degree polynomials over sliding windows to preserve trends and peaks capturing trends. It provides a smoothed estimate of the mean arrival rate within each traffic phase, which helps to reveal state-dependent behavior in MMPP.

The preprocessing pipeline contains several hyperparameters, including the EMA smoothing factor ($\alpha$), the Savitzky--Golay window length and polynomial order, and the moving-average window used for estimating the global arrival rate. To determine an effective configuration, we conducted a sensitivity analysis by varying each parameter over a suitable range while keeping the remaining parameters fixed. For each parameter configuration, we performed three independent arrival-rate estimation experiments using MMPP-2 models with state arrival rates of (50,100), (75,175), and (100,250), thereby evaluating the estimator across progressively larger differences in traffic intensity. Since the underlying MMPP-2 process provides the true arrival rate for every request, the quality of the online estimator was assessed by comparing the estimated arrival rate against the corresponding ground-truth state arrival rate. Estimation accuracy was quantified using the Mean Absolute Error (MAE) and Mean Absolute Percentage Error (MAPE), computed over the entire request trace for each scenario. The average MAE and MAPE across all three scenarios were then used to rank parameter configurations, ensuring that the selected preprocessing pipeline remained robust across varying traffic intensities and burst magnitudes. The configuration with the lowest average estimation error was adopted for all subsequent experiments. The optimal setting was found to be $\alpha$ = 0.05, Savitzky--Golay window length = 151, polynomial order = 2, and moving-average window = 5, achieving an average MAE of 15.39 and an average MAPE of 14.37\%, with maximum MAE and MAPE of 17.78 and 16.13\%, respectively.

\begin{table}[H]
\centering

\label{tab:sensitivity}
\small
\begin{tabular}{|c|c|c|c|}

\multicolumn{4}{c}{\textbf{Hyperparameter Configuration}}\\
\hline
EMA ($\alpha$) & SG Window & SG Polynomial & MA Window \\
\hline
0.05 & 151 & 2 & 5 \\
\hline
\end{tabular}
\caption{Optimal parameter configuration}
\end{table}








\begin{figure}
    \centering
    \includegraphics[width=0.85\linewidth]{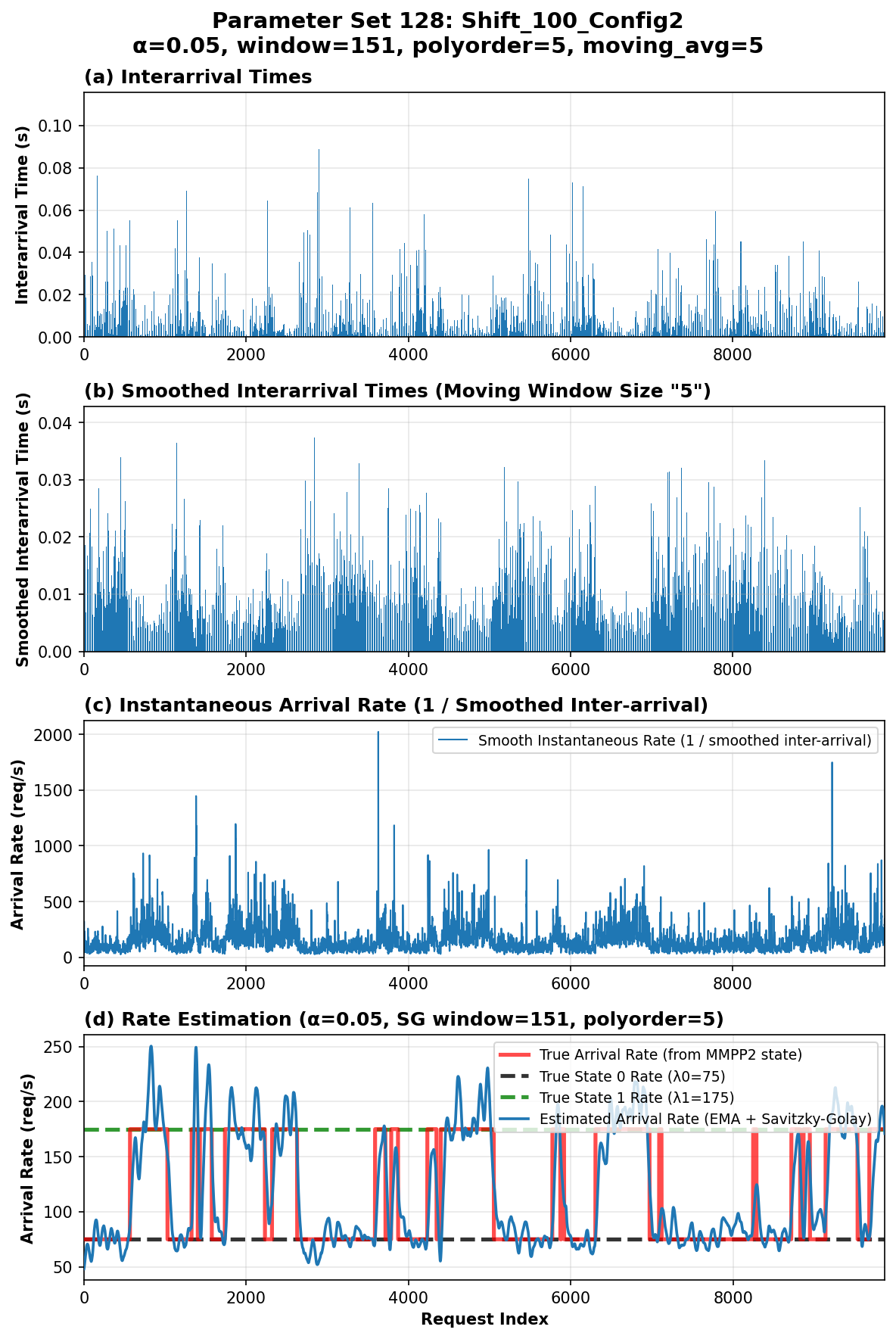}
    \caption{Preprocessing of Request Arrival Trace}
    \label{fig:preprocessing}
\end{figure}

\begin{figure}
    \centering
    \includegraphics[width=0.4\linewidth]{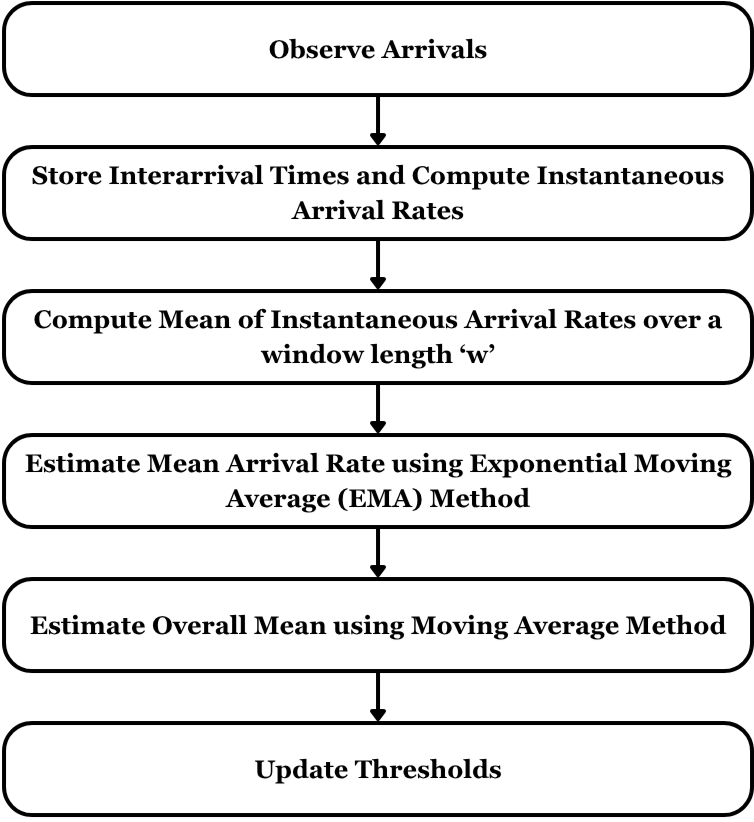}
    \caption{Pipeline for Estimating Arrival Rates and Updating Thresholds}
    \label{fig:pipeline_preprocessing}
\end{figure}

We also maintain a moving average over all EMA values to estimate a stable global mean arrival rate. This global rate is then used to dynamically adapt thresholds in the WAIT scheduling algorithm, enabling responsive batching under traffic fluctuations. Figure~\ref{fig:preprocessing} shows our preprocessing pipeline applied to an MMPP-2 request trace with state arrival rates of 175 and 75. The figure uses the best-performing set of preprocessing parameters obtained from the sensitivity analysis, i.e., the configuration with the lowest average MAE and MAPE across all evaluation scenarios.Figure 4(a) and 4(b) plot interarrival and mean interarrival times over a smooth window while figure 4 (c) plots the smooth instantaneous arrival rate against request index. In Figure 4 (d) smooth instantaneous rate, actual arrival rate and estimated arrival rate are plotted w.r.t request index. Figure~\ref{fig:pipeline_preprocessing} outlines the process starting from the arrival of requests to the modification of thresholds.The adaptive threshold update in the Modified WAIT algorithm follows a feedback mechanism driven by the online arrival rate estimate $\hat{\lambda}_t$, obtained via the exponential moving average (EMA) process. In Equation~(1), the batching threshold $n_j$ is inversely related to the mean arrival rate $\lambda_j$. Therefore, replacing $\lambda_j$ with the time varying estimate $\hat{\lambda}_t$ yields a dynamic threshold
\[
n_j(t) \propto \frac{1}{\hat{\lambda}_t},
\]
which decreases during traffic surges, enabling faster batch formation, and increases during low load intervals, preventing GPU underutilization and cache churn. This adaptive behavior effectively smooths throughput fluctuations, maintaining near optimal batching efficiency under bursty and non stationary workloads.

\subsection{Estimating Arrival Rates: Multiple Request Types}
\begin{figure}
    \centering
    \includegraphics[width=0.6\linewidth]{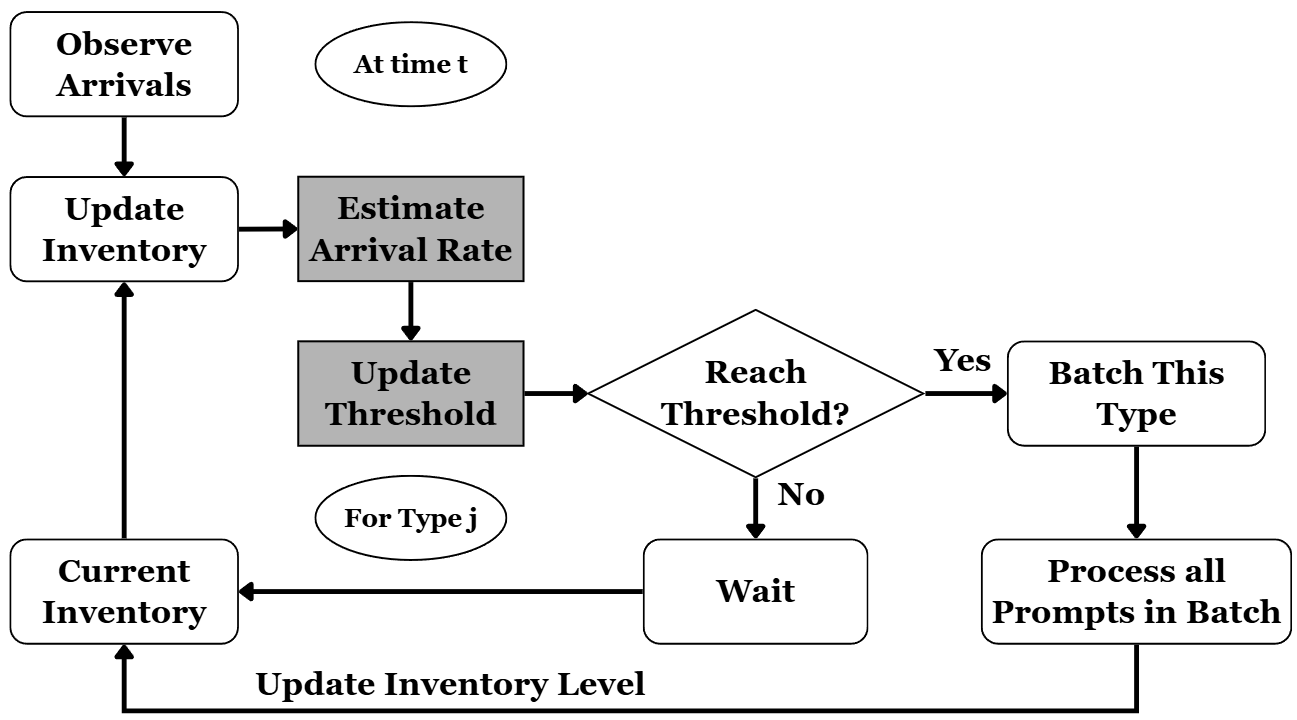}
    \caption{Pipeline of the Modified WAIT Algorithm. The grey action blocks represent our modification to the existing WAIT Algorithm.}
    \label{fig:flowchart_wait}
\end{figure}

In scenarios with multiple request types, we individually estimate the arrival rate for each type using our method. These per-type arrival rate estimates are then used to compute the thresholds for the WAIT scheduling algorithm. Figure~\ref{fig:flowchart_wait} shows the control flow of the Modified WAIT algorithm extended to multiple request types. Each type $j \in \{1,\ldots,m\}$ maintains its own arrival rate estimator, inventory counter, and adaptive threshold $n_j(t)$, allowing heterogeneous prompt classes to evolve independently. As new prompts arrive, their inter-arrival times are smoothed via the exponential moving average (EMA) filter (Section~V-B) to obtain $\hat{\lambda}_j(t)$, which feeds into the \emph{Estimate Arrival Rate} block. The subsequent \emph{Update Threshold} block applies the dynamic rule
\[
n_j(t) \propto \frac{1}{\hat{\lambda}_j(t)},
\]
thereby adjusting batching thresholds in real time.

The algorithm then identifies the prompt types whose inventories meet their thresholds and forms batches of $\min\!\big(n_j(t),\, n_{jk}^t\big)$ prompts across processing stages $k$. Requests below the threshold continue to accumulate while retaining their KV caches. This per-type adaptive loop balances GPU utilization across heterogeneous workloads by rapidly batching high-frequency prompt types and deferring low-frequency ones, thereby creating a self-correcting feedback mechanism that maintains stable throughput under multi-tenant, bursty traffic conditions.

\subsubsection{Intuition Behind Adaptive Thresholding}

The original WAIT algorithm assumes that arrival rates are known a priori and remain constant throughout execution. Under bursty workloads, however, the effective arrival rate changes continuously as the workload transitions between different traffic regimes. Consequently, batching thresholds computed from a fixed arrival rate gradually become inconsistent with the current workload, leading to either premature batch formation or unnecessary waiting.

The proposed approach addresses this limitation by continuously estimating the arrival rate from the observed request stream and updating the batching thresholds accordingly. Rather than relying on a single workload estimate throughout execution, the scheduler adapts its batching decisions using the most recent traffic information. This enables the scheduler to respond to workload fluctuations while preserving the original threshold-based scheduling framework of WAIT.

\section{Numerical Experiments}

This section presents a comprehensive empirical evaluation of various scheduling algorithms for LLM inference. We assess their performance on synthetically generated workloads. The synthetic datasets are created by simulating request arrivals using the MMPP, specifically the two-state variant (MMPP-2). A detailed description of the synthetic trace generation methodology was provided in earlier sections.

All experiments are conducted using Microsoft Vidur \cite{2405.05465} to simulate an NVIDIA A100 GPU, mirroring the hardware configuration employed in prior studies \cite{agrawal2024sarathi}. We benchmark our algorithm against vLLM, ORCA and Sarathi, widely recognized inference engines that employ FCFS scheduling. Specifically, vLLM prioritizes new arrivals, while Sarathi prioritizes ongoing prompts, with both enforcing fixed limits on the total number of tokens and prompts per iteration. We also include the performance of the Original WAIT algorithm with knowledge of the entire request arrival trace. Specifically, we execute the WAIT algorithm using thresholds derived from the overall mean arrival rate corresponding to each request type.  To ensure a fair comparison, we enforce the same batch size limit across all the algorithms.

We test the four algorithms under four scenarios:
\begin{enumerate}
    \renewcommand{\labelenumi}{\roman{enumi}.}
    \item Low Demand Low Shift (LDLS)
    \item Low Demand High Shift (LDHS)
    \item High Demand Low Shift (HDLS)
    \item High Demand High Shift (HDHS)
\end{enumerate}
Low Demand and High Demand correspond to request arrival traces with low and high arrival rates respectively. Similarly, Low Shift and High Shift correspond the request arrival traces with low and high transition rates between different states of requests.

\subsection{Low Demand}
`Low Demand' experiments feature comparatively low arrival rates. The numerical values used to simulate LDLS and LDHS scenarios are tabulated in Table~\ref{tab:simulation_params}.

\renewcommand{\arraystretch}{1.2}
\begin{table}
\centering
\caption{Simulation Parameters for Low Demand Scenarios}
\label{tab:simulation_params}

\begin{tabular}{|p{3cm}|p{3cm}|}
\hline
\textbf{Parameter} & \textbf{LDLS \& LDHS} \\
\hline
$m$ & $2$ \\
\hline
($l_1, l_2$) & (10, 10) \\
\hline
($l'_1, l'_2$) & (10, 20) \\
\hline
($\lambda_{1,0}, \lambda_{2,0}$) & (1000, 1000) \\
\hline
($\lambda_{1,1}, \lambda_{2,1}$) & (500, 200) \\
\hline
\end{tabular}

\vspace{6pt}

\begin{tabular}{|p{3cm}|p{1.3cm}|p{1.3cm}|}
\hline
\textbf{Parameter} & \textbf{LDLS} & \textbf{LDHS} \\
\hline
$q^m_{0,1}$ $\forall m \in \{1, 2\}$ & $0.5$ & $1.5$ \\
\hline
$q^m_{1,0}$ $\forall m \in \{1, 2\}$ & $0.3$ & $1.3$ \\
\hline
\end{tabular}
\end{table}
\renewcommand{\arraystretch}{1.0}

where $m$ refers to the number of request types, $l_1, l_2$ refer to the prefill tokens of the two types, $l'_1 l'_2$ refer to the number of decode tokens. $\lambda_{1,0}$ and $\lambda_{2,0}$ refer to the mean arrival rates of request types 1 and 2 in state 0, respectively. Similarly, $\lambda_{1,1}$ and $\lambda_{2,1}$ refer to the mean arrival rates of requests types 1 and 2 in state 1. $q_{0,1}^m$ and $q_{1,0}^m$ are a quantifier of the rate of transition between different states of request type $m$. These notations remain consistent throughout the subsequent sections.
 
\begin{figure*}
\centering
\includegraphics[width=.5\textwidth]{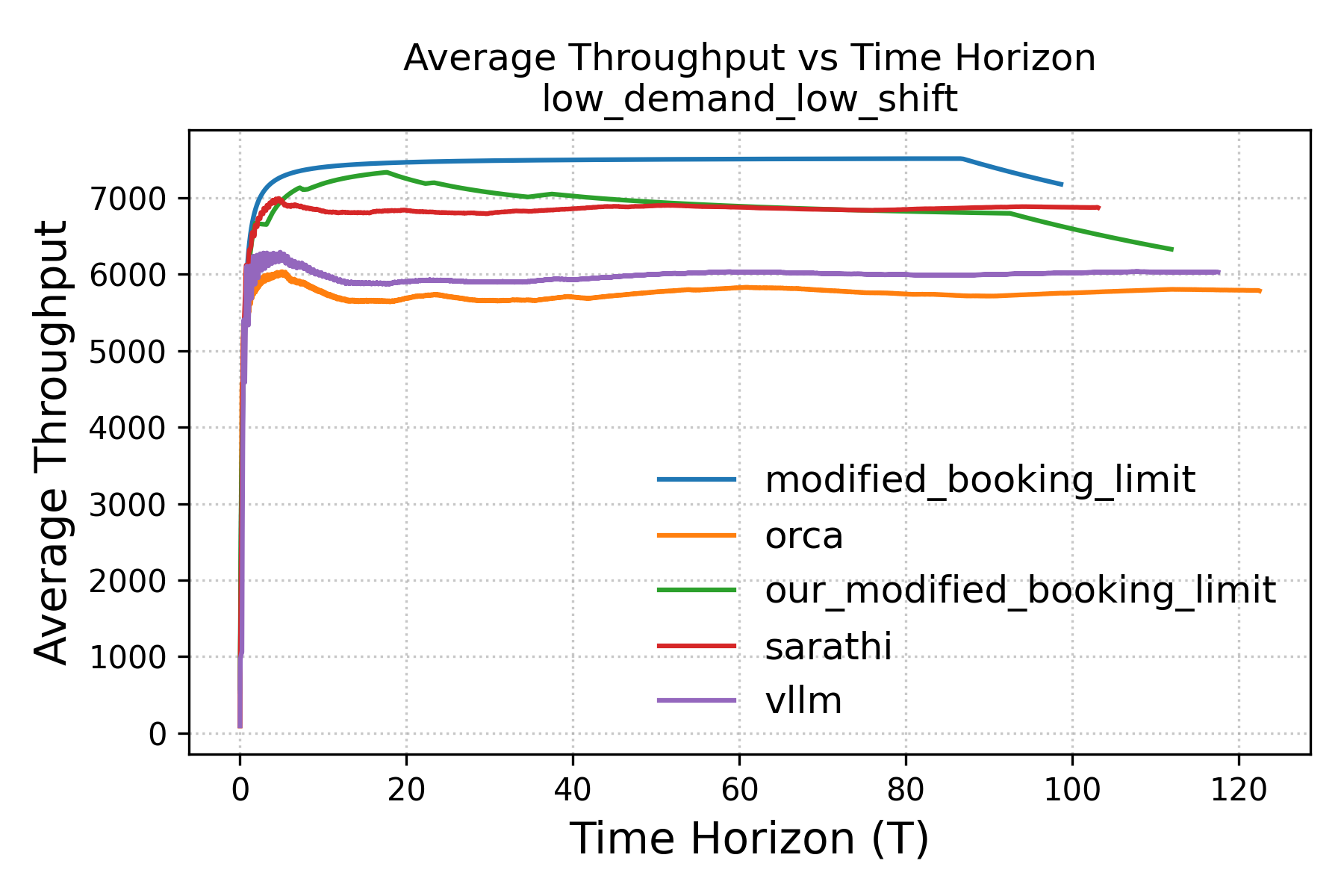}\hfill
    \includegraphics[width=.5\textwidth]{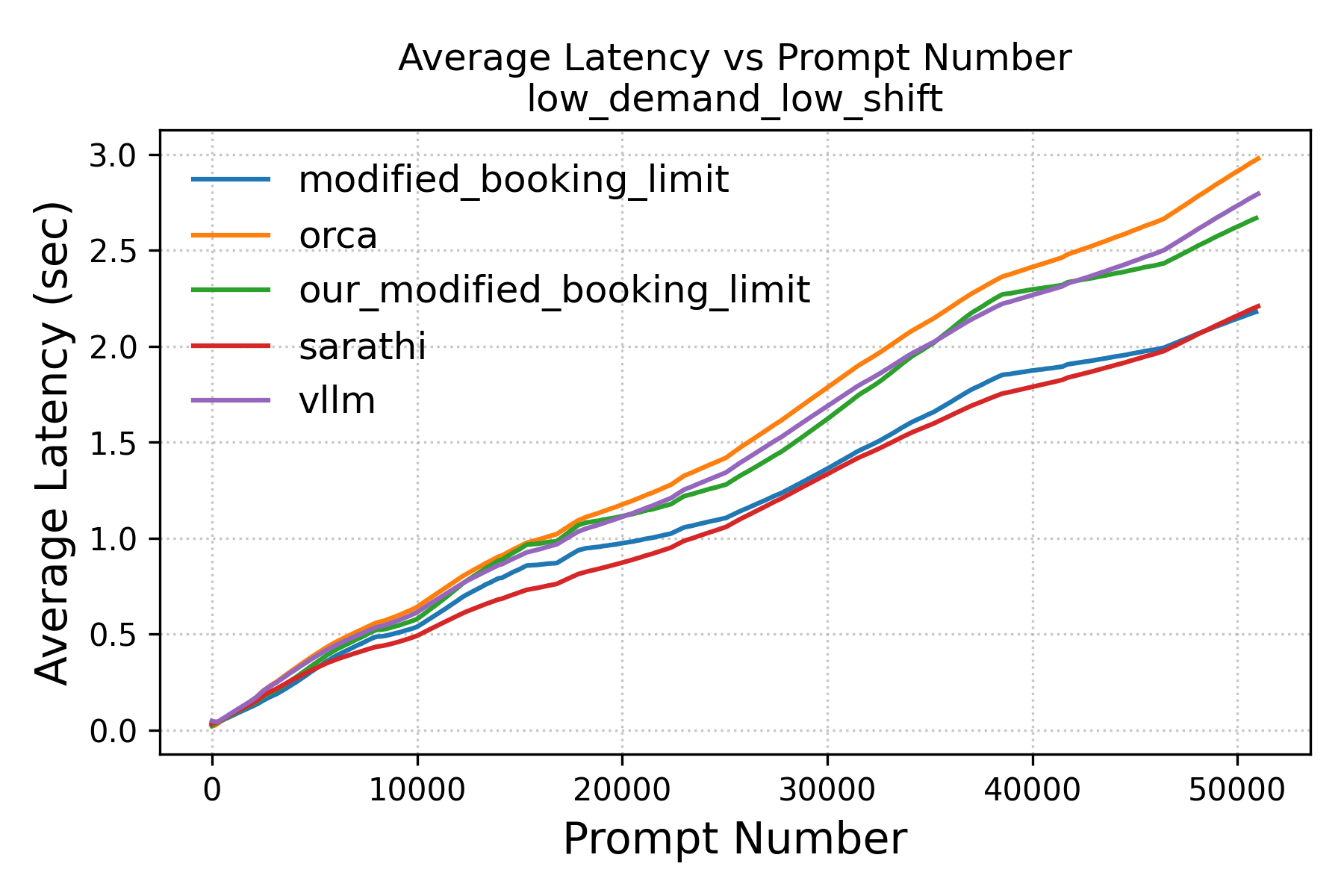}\hfill
    \\
    \includegraphics[width=.5\textwidth]{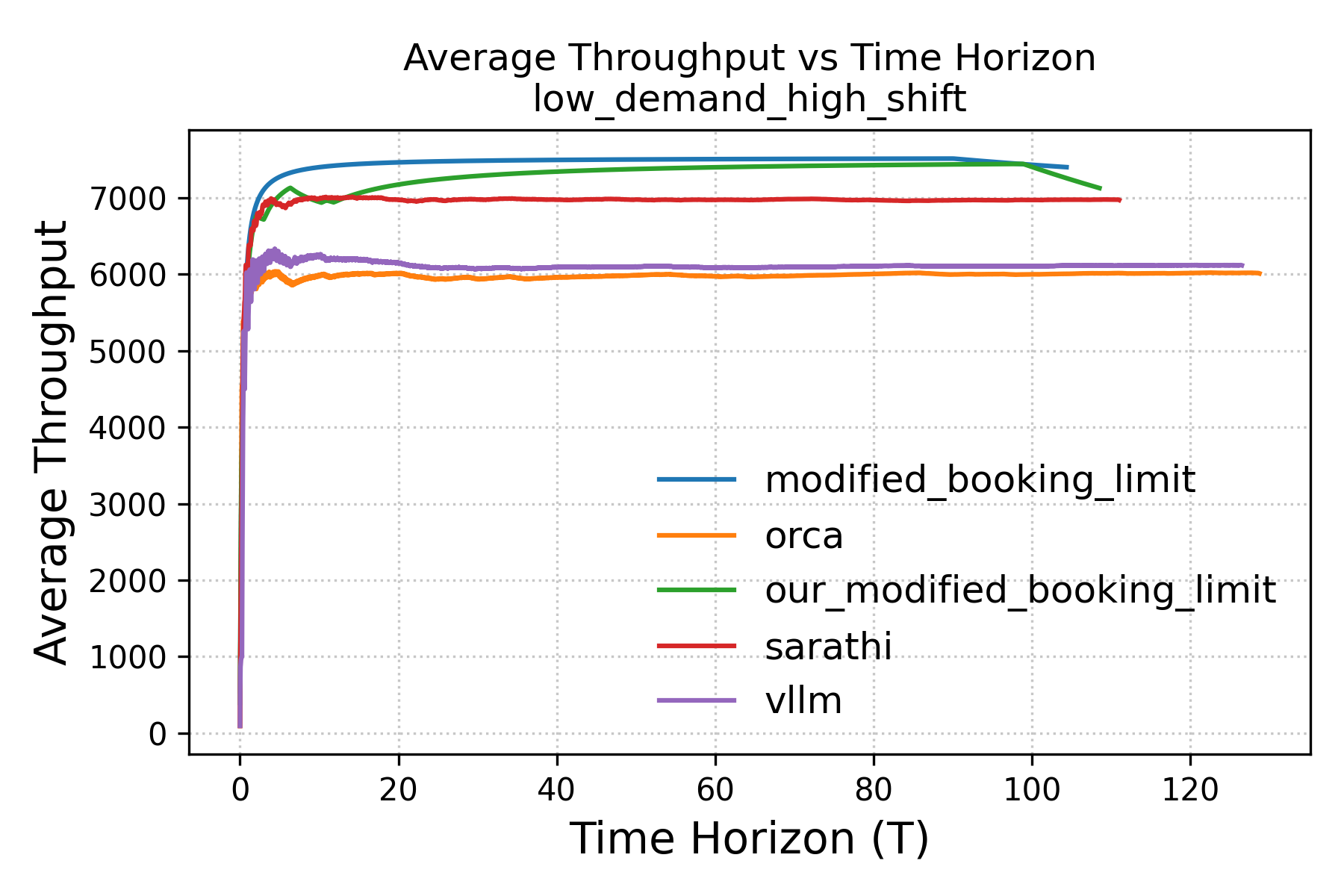}\hfill
    \includegraphics[width=.5\textwidth]{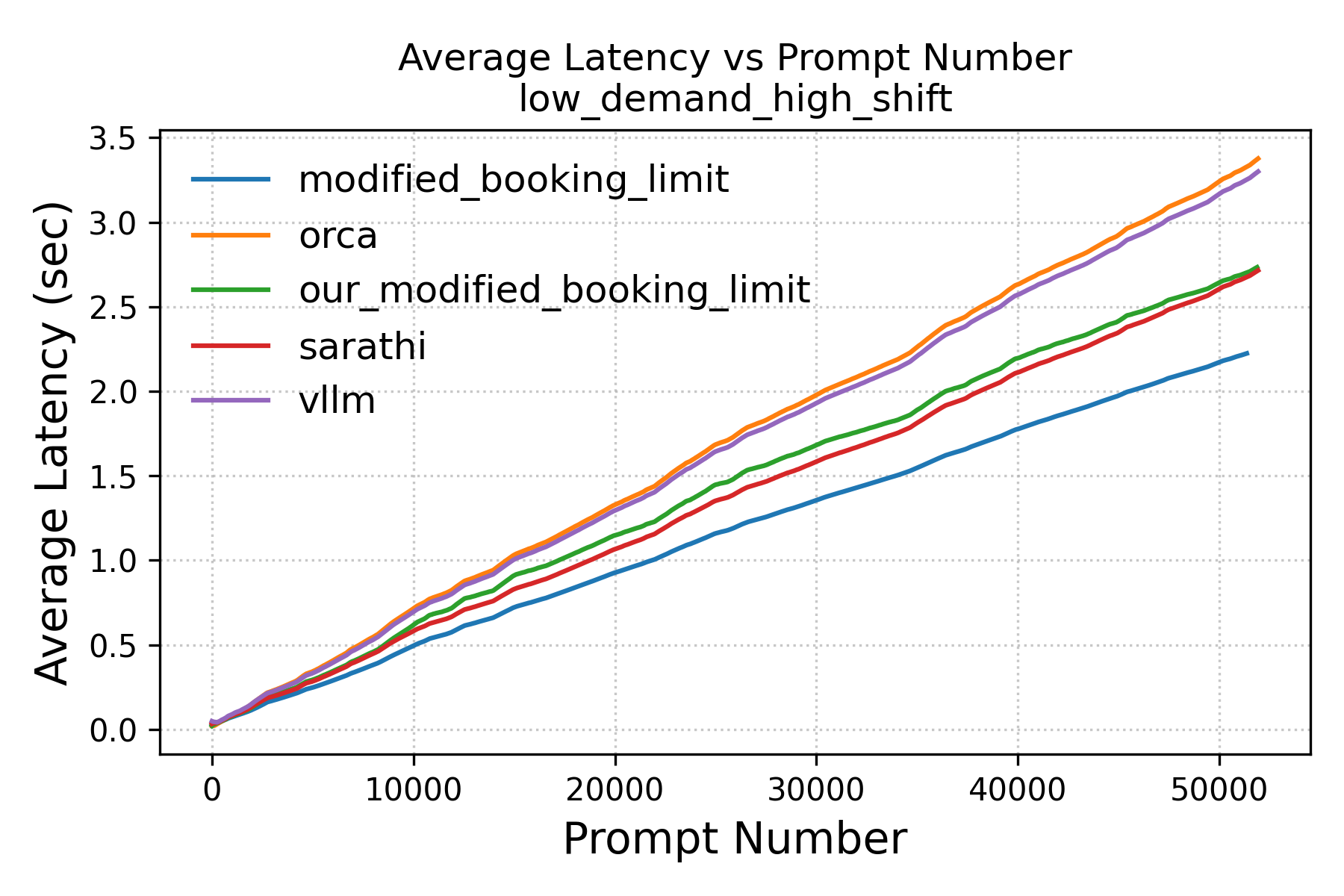}\hfill
    \caption{Average throughput and latency on datasets with Low Demand}\label{fig:low_demand_results}
\end{figure*}

In Figure~\ref{fig:low_demand_results}, we present the average throughput and latency results under low-demand scenarios as categorized into (i) LDLS and (ii) LDHS. In both figures, the red line denotes the performance of the WAIT algorithm, which assumes prior knowledge of the request arrival distribution. This represents the best possible performance under ideal conditions. However, such prior knowledge is typically unavailable in real-world settings. Despite this limitation, our Modified-WAIT 

\renewcommand{\arraystretch}{1.2}
\begin{table}
\centering
\caption{Experiment Parameters for high demand configurations}
\begin{tabular}{|p{3.5cm}|p{3.5cm}|}
\hline
\textbf{Parameter} & \textbf{HDHS \& HDLS} \\
\hline
$m$ & 3 \\
\hline
($l_1, l_2, l_3$) & (20, 20, 20)\\
\hline
($l'_1, l'_2, l'_3$) & (100, 200, 300) \\
\hline
($\lambda_{1,0}, \lambda_{2,0}, \lambda_{3,0}$) & (5000, 4000, 2000) \\
\hline
($\lambda_{1,1}, \lambda_{2,1}, \lambda_{3,1}$) & (4000, 3000, 1000) \\
\hline
\end{tabular}

\vspace{6pt}

\begin{tabular}{|p{4cm}|p{1.3cm}|p{1.3cm}|}
\hline
\textbf{Parameter} & \textbf{HDHS} & \textbf{HDLS} \\
\hline
$q^m_{0,1}$ $\forall m \in \{1, 2, 3\}$ & 0.5 & 1.5 \\
\hline
$q^m_{1,0}$ $\forall m \in \{1, 2, 3\}$ & 0.3 & 1.3 \\
\hline
\end{tabular}

\label{tab:hdls_hdhs}
\end{table}
\renewcommand{\arraystretch}{1.0}

algorithm achieves performance that closely approximates that of WAIT, highlighting its robustness even without access to arrival distribution information.

In terms of latency, our algorithm exhibits performance on par with WAIT under high-shift scenarios, outperforming other algorithms. In low-shift settings, while it does not outperform all four baselines, it still achieves latency performance that remains closely aligned with the original WAIT algorithm.


\begin{figure*}
\centering
    \includegraphics[width=.5\textwidth]{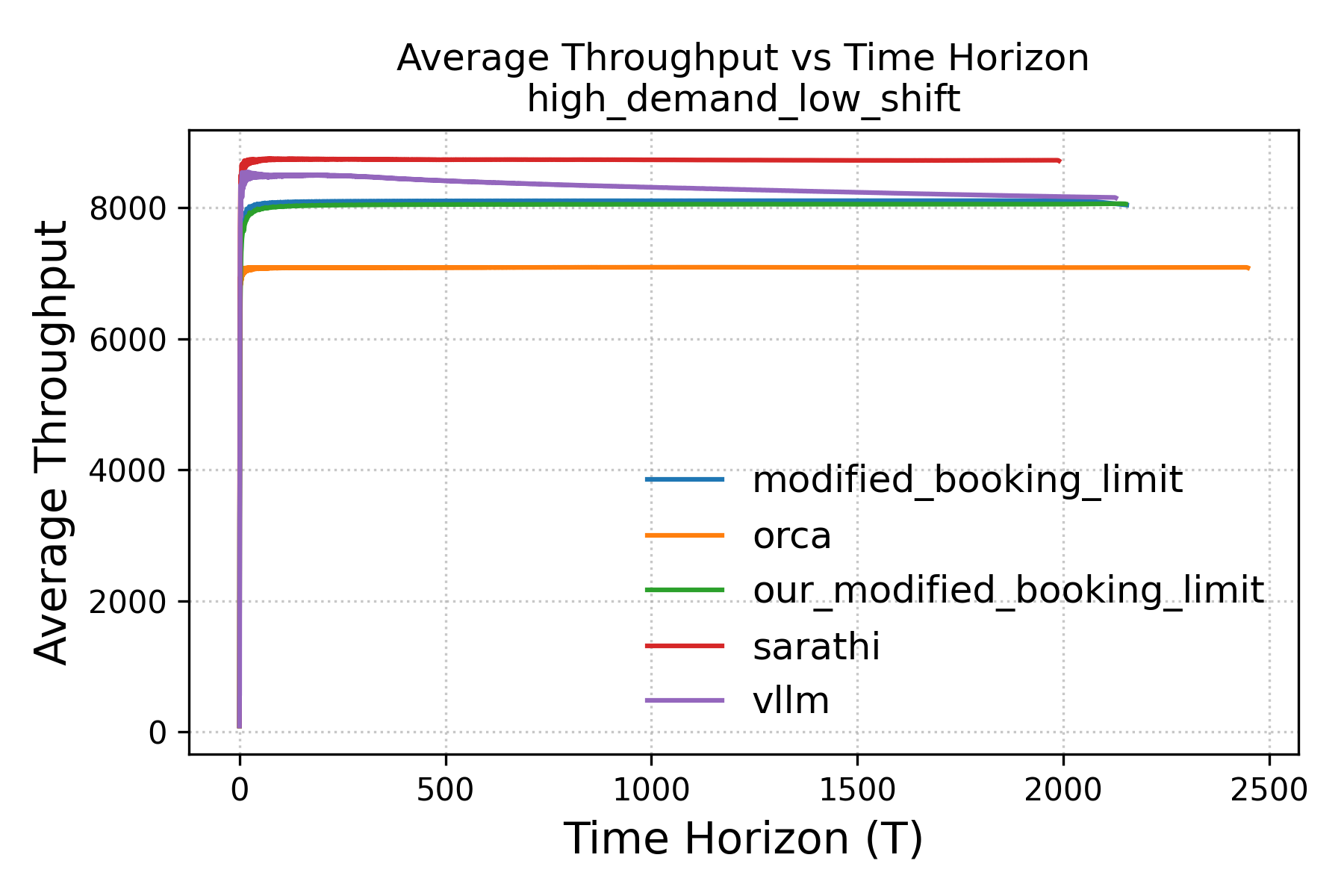}\hfill
    \includegraphics[width=.47\textwidth]{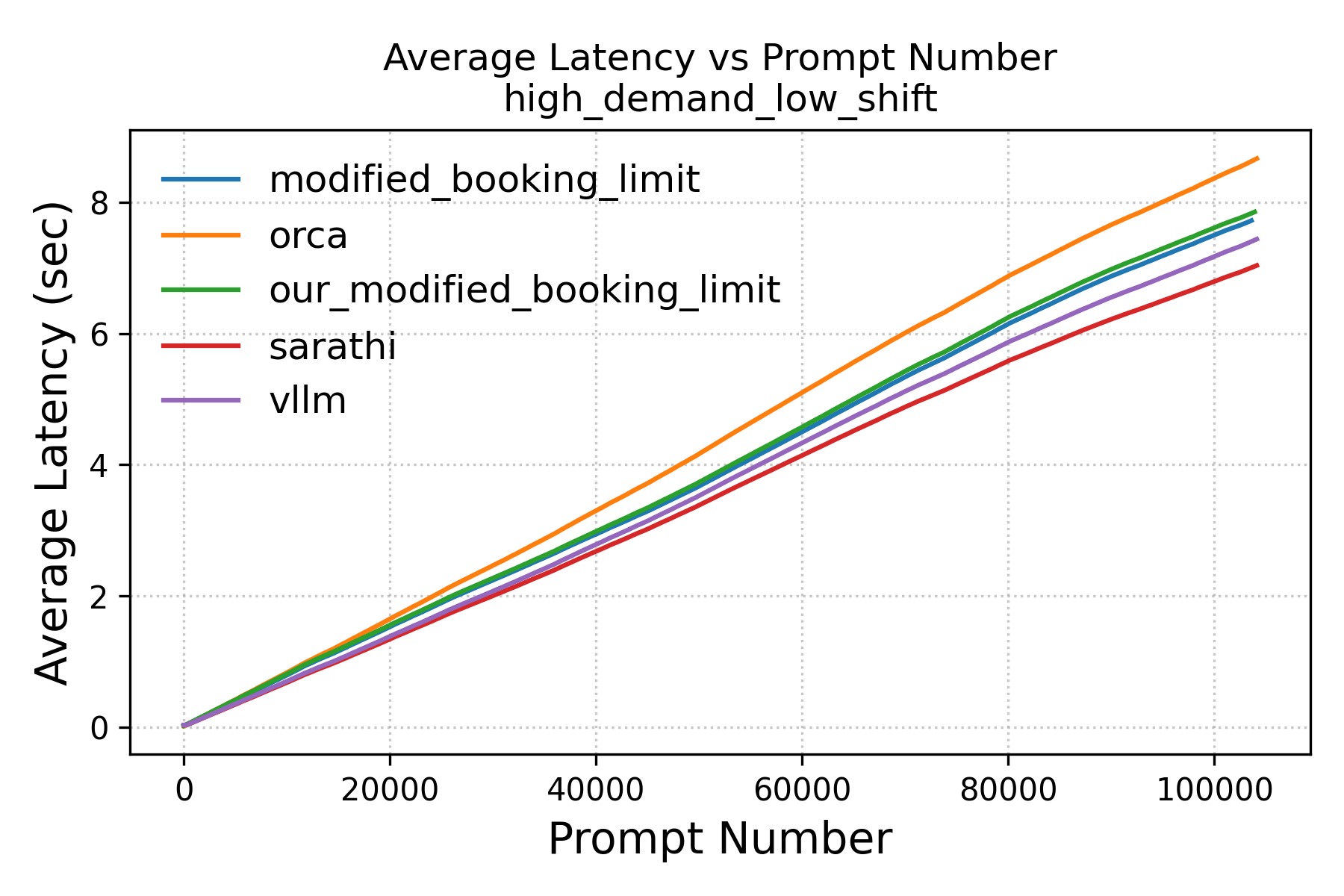}\hfill
    \\
    \includegraphics[width=.5\textwidth]{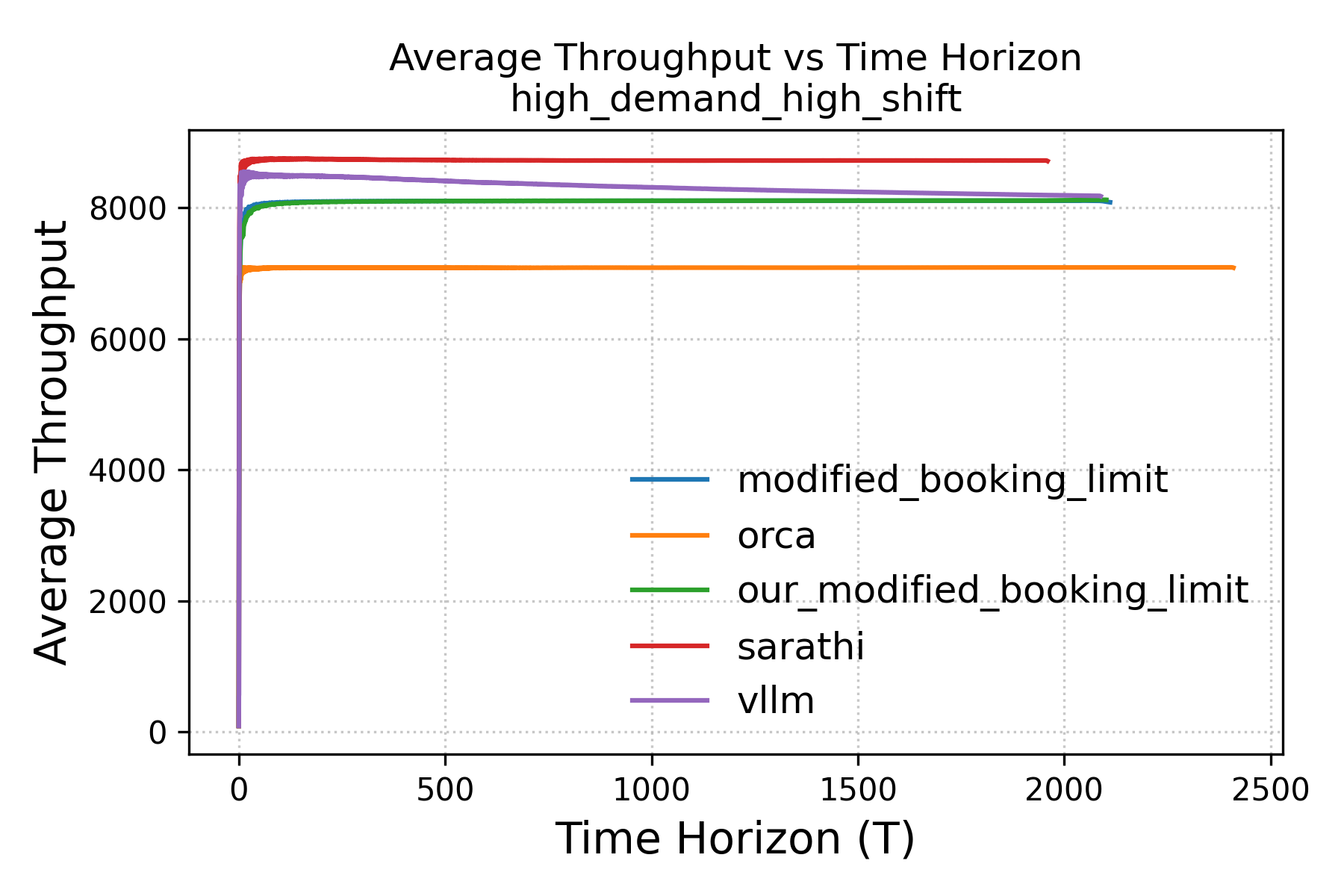}\hfill
    \includegraphics[width=.475\textwidth]{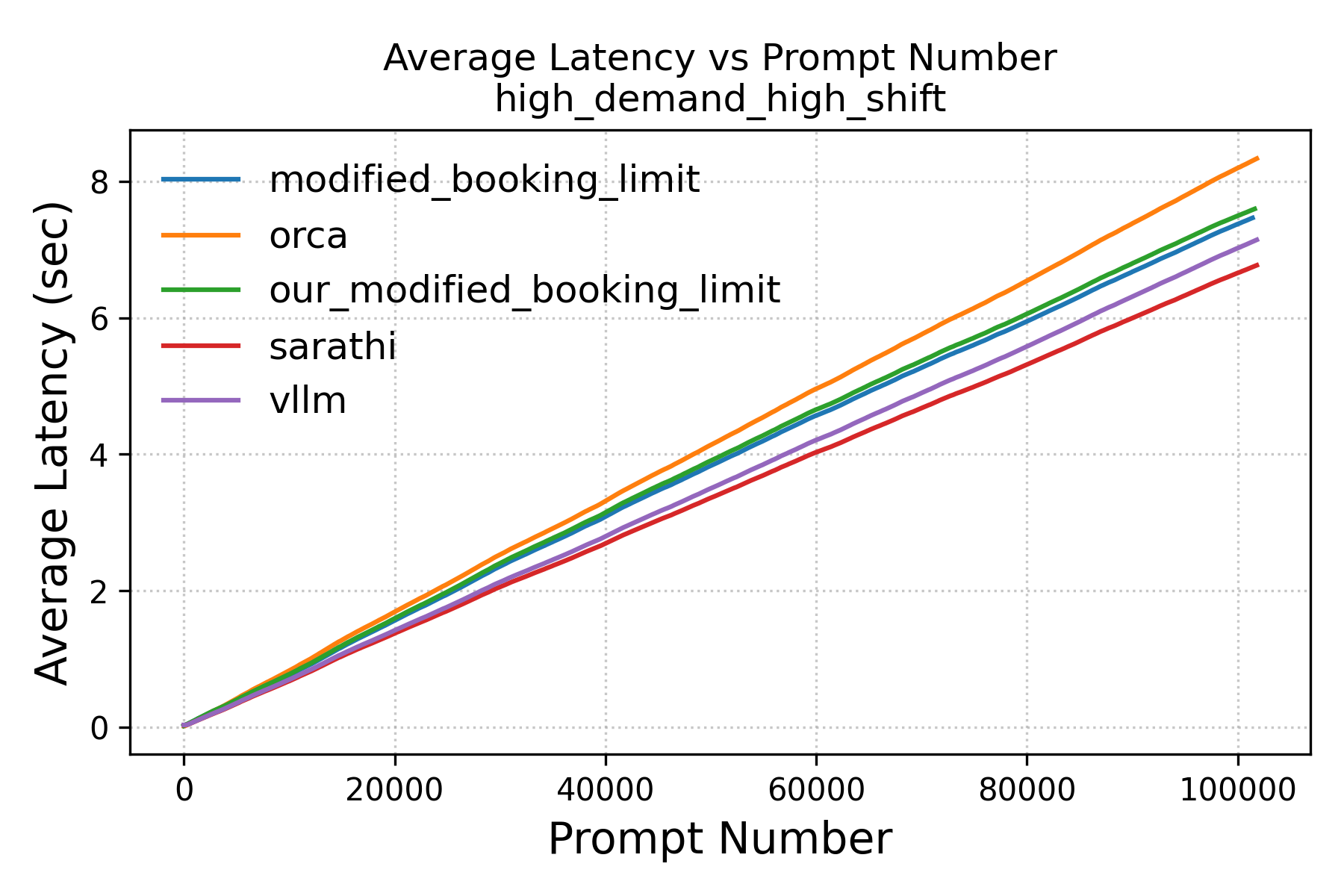}\hfill
    \caption{Average throughput and latency on datasets with High Demand}\label{fig:high_demand_results}
\end{figure*}

\subsection{High Demand}
'High Demand' experiments feature comparatively high arrival rates. The experiment parameters used to simulate HDLS and HDHS scenarios are detailed in Table~\ref{tab:hdls_hdhs}.

Similarly, Figure~\ref{fig:high_demand_results} presents the average throughput and average latency results under high-demand scenarios. In these subcategories, our Modified-WAIT algorithm performs closely to the original WAIT algorithm in terms of average throughput for both HDLS and HDHS. However, in terms of latency, it notably surpasses the original WAIT algorithm.



In summary, our Modified WAIT algorithm performs is comparable to the original WAIT algorithm across all test cases. Furthermore our model outperforms both Sarathi and vLLM in terms of throughput under `Low Shift' scenarios. Under `High Shift' scenarios both original WAIT and modified WAIT need improvements for better performance. While latency remains suboptimal across all settings for both versions of WAIT, performance improves significantly in the `Low Shift' regime.

\section{Conclusion}
Our study focuses on LLM inference under bursty workload conditions. To evaluate scheduling performance under bursty traffic patterns, we simulate request arrivals using an MMPP-2. We compare our proposed model with three existing  scheduling algorithms: WAIT, Sarathi, and vLLM, and the corresponding findings are listed below:

\begin{itemize}
    \item We first determine the optimal thresholds for the WAIT algorithm, assuming that the request arrival distribution is known.
    \item To approach this ideal performance in practical settings where distributions are unknown, we developed a Modified WAIT algorithm.
    \item Our Modified WAIT approach achieves throughput and latency close to the ideal WAIT with prior knowledge.
    \item It significantly outperforms Sarathi and vLLM in throughput, especially under 'Low Shift' burst conditions.
\end{itemize}

Our findings highlight the importance of workload-aware scheduling and pave the way for more robust LLM serving strategies in production environments. While achieving promising results, there are several directions for future improvement:

\begin{itemize}
    \item Integrating the EMA method with Changepoint Detection to accurately estimate the mean arrival rates of each state.
    \item While our simulations using Microsoft Vidur produce results that closely reflect reality, future work could involve validating our algorithms on real-world datasets consisting of actual prompts and responses.
    \item Another promising direction is developing a machine learning model that predicts the type of incoming requests at arrival time. Such a model would make the Modified WAIT algorithm more applicable to real-world deployment scenarios.
\end{itemize}


%



\section*{Acknowledgment}
The first author (A.G.Katageri) would like to thank the Indian Institute of Technology Kanpur for organizing the SURGE Internship Program. The second author (S.Rani) would like to thank the Indian Institute of Technology Kanpur for providing the research environment and support through the Institute Postdoctoral Fellowship.




\bibliographystyle{elsarticle-num}
\bibliography{references}
%

%

\end{document}